\documentclass[a4paper,oneside,12pt,english]{article}
\usepackage{url}
\usepackage[pdftex]{color,graphicx}
\usepackage{hyperref}
\hypersetup{plainpages = false,
              breaklinks=true,
              linktocpage,
              colorlinks   = true, 
              urlcolor     = blue, 
              linkcolor    = blue, 
              citecolor   = red, 
              pdfauthor={Niko Brummer},
              pdfstartpage=1, 
              pdfstartview=FitH,  
              pdfkeywords={}}

\usepackage{amsmath}%
\usepackage{amsfonts}%
\usepackage{amssymb}%
\usepackage{graphicx}

\usepackage{tikz}
\usetikzlibrary{bayesnet}

\def\gammavec{\boldsymbol{\gamma}}

\def\phivec{\boldsymbol{\phi}}

\def\muvec{\boldsymbol{\mu}}

\def\Wmat{\mathbf{W}}
\def\Cmat{\mathbf{C}}
\def\Tmat{\mathbf{T}}

\def\Bmat{\mathbf{B}}
\def\Mmat{\mathbf{M}}

\def\avec{\mathbf{a}}
\def\xvec{\mathbf{x}}
\def\mvec{\mathbf{m}}
\def\fvec{\mathbf{f}}

\def\Imat{\mathbf{I}}

\def\ND{\mathcal{N}}
\def\R{\mathbb{R}}

\def\Sset{\mathcal{S}}

\def\LB{\mathcal{L}}

\def\const{\texttt{const}}

\DeclareMathOperator{\trace}{tr}

\def\expv#1#2{\left\langle#1\right\rangle_{#2}}

\def\logdetm#1{\log\bigl\lvert#1\bigr\rvert}

\begin{document}
\title{Language-depedent I-Vectors for LRE15}
\author{Niko Br\"ummer and Albert Swart\\ AGNITIO Research South Africa}
\date{Brno, October 2015}
\maketitle

\abstract{A standard recipe for spoken language recognition is to apply a Gaussian back-end to i-vectors. This ignores the uncertainty in the i-vector extraction, which could be important especially for short utterances. A recent paper by Cumani, Plchot and Fer proposes a solution to propagate that uncertainty into the backend. We propose an alternative method of propagating the uncertainty.}

\section{Introduction}
A standard recipe for language recognition via i-vectors~\cite{LRivec}, is to extract the i-vectors---i.e.\ \emph{point estimates} of the hidden variables---and then score them using a linear Gaussian back-end (LGBE).\footnote{The LGBE has a common, within-class covariance, shared by all languages and language-dependent means. The score is linear (affine, from i-vector to score vector), because the quadratic term in the Gaussian log-likelihood is language-independent and cancels.}

In this document, we combine the LGBE and the i-vector model into one monolithic model. In training and test, we can now \emph{integrate} out the hidden i-vectors to directly produce language recognition scores, without having to go via explicit point estimates of the i-vectors. 

Since the i-vector model is intractable in closed form, we resort to mean-field VB, using an approximate posterior where the GMM state path and i-vector posterior are independent. We compare our recipe to a similar one by Cumani, Plchot and F\'er~\cite{CPF}, which makes use of a language-independent i-vector posterior approximation. In our recipe, the i-vector posterior approximation is instead language-dependent and can be expected to more closely approximate the true posterior.

On a practical note, if we already have extracted i-vectors, we can still apply our scoring recipe, as long as we have available the zero-order stats associated with each i-vector.

\section{The model}
\begin{figure}[!htb]
\centerline{
\begin{tikzpicture}
\node[latent] (ivec) {$\xvec_{s}$};
\node[const, outer sep = 15pt] at(ivec|-ivec) (dummy1) {};
\node[obs, right=of ivec] (data) {$\phivec_{st}$};
\node[const, outer sep = 15pt] at(data|-data) (dummy2) {};
\node[latent, right=of data] (state) {$\gammavec_{st}$};
\node[const, outer sep=4pt, above=40pt of data] (T) {$\Tmat$};
\node[const, outer sep=4pt, left=of ivec] (prior) {$\ND(\mvec_\ell,\Wmat^{-1})$};
\plate[draw=blue,inner sep = 5pt] {innerplate} {(data)(state)(dummy2)} {$t\in\{1,\ldots,T_s\}$};
\plate[draw=blue,inner sep = 5pt] {} {(innerplate)(ivec)(dummy1)} {$s\in\Sset_\ell$};
\edge {ivec}{data};
\edge {prior}{ivec};
\edge {state}{data};
\edge {T}{data};
\end{tikzpicture}
}
\caption{\textbf{The i-vector model.} Language, segment and frame are indexed with $\ell,s,t$ and $\Sset_\ell$ is the set of all indices, $s$, for which the language is $\ell\in\{1,\ldots,L\}$. The hidden i-vector is $\xvec_{s}$, the hidden GMM state is $\gammavec_{st}$ and the observed feature vector is $\phivec_{st}\in\R^D$. The factor loading matrix is $\Tmat$. The UBM parameters are not shown.} 
\label{fig:model}
\end{figure}
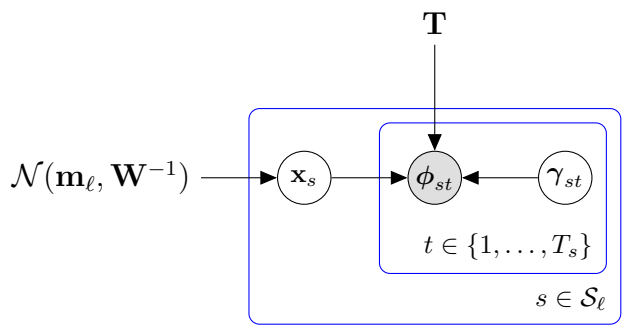

\def\philt{\phivec_{{\ell}t}}
The model is shown in figure~\ref{fig:model}. It is almost a standard i-vector extractor, except that we have allowed the i-vector prior to be non-standard, with a language dependent mean, $\mvec_\ell$ and a within-class precision, $\Wmat$. 

\noindent This model incorporates the i-vector extractor and a linear Gaussian back-end into one and will allow joint training of both and will allow language scores to be directly extracted, without having to go via intermediate i-vectors.

\section{Plugin model parameters}
Here we derive a plugin ML recipe for the model parameters $\Wmat$ and the $\muvec_\ell$. (The i-vectors, $\xvec_s$, are not plugged in, but instead integrated out.)

As mentioned in the introduction, we make the model tractable by a mean-field VB approach. More details of this approach can be found in~\cite{VBIvector}.

\subsection{The VB lower bound}
We handle this model via mean-field VB, where the approximate posterior for the GMM path is fixed and given by the UBM \emph{responsibilities}, $q_{st}^i$, which sum to unity over states, $i$. The i-vector posterior, $Q_{\ell s}(\xvec)$ is language-dependent. The VB lower bound thus obtained is: 
\begin{align}
\label{eq:VBLE}
\LB_{\ell s} &= \expv{\log\frac{\ND(\xvec\mid\mvec_\ell,\Wmat^{-1})}{Q_{\ell s}(\xvec)}
+\sum_{t=1}^{T_s} \sum_{i=1}^N q_{st}^i \log\frac
{w_i\ND(\phivec_{st}^i\mid\Tmat_i\xvec,\Imat)}
{q_{st}^i} }{Q_{\ell s}} 
\end{align}
where $\phivec_{st}^i$ denotes a feature vector, centred and whitened w.r.t.\ the parameters of UBM component $i$, so that we can further ignore the UBM parameters. 

\subsection{The i-vector posterior}
\label{sec:Qls}
The approximate i-vector posterior is:
\begin{align}
\label{eq:mfivecQ2}
\begin{split}
\log Q_{\ell s}(\xvec)
&= \log \ND(\xvec\mid\mvec_\ell,\Wmat^{-1}) + \sum_{t=1}^{T_{s}} \sum_{i=1}^N q_{st}^i \log \ND(\phivec_{st}^i\mid \Tmat_i\xvec,\Imat) + \const \\
&= \xvec'\left[\Wmat\mvec_\ell + \sum_i\Tmat'_i \Bigl(\sum_{t} q_{st}^i \phivec_{st}^i\Bigr)\right]
-\frac12\xvec'\left[\Wmat+\sum_i \Tmat'_i\Tmat_i \Bigl(\sum_t q_{st}^i\Bigr) \right]\xvec
+\const\\
&= \xvec'\left[\Wmat\mvec_\ell + \sum_i\Tmat'_i \fvec_{s}^i\right]
-\frac12\xvec'\left[\Wmat+\sum_i \Tmat'_i\Tmat_i n_{s}^i \right]\xvec
+\const\\
&= \xvec'\left[\Wmat\mvec_\ell + \avec_s \right]
-\frac12\xvec'\left[\Wmat+ \Bmat_s \right]\xvec
+\const
\end{split}
\end{align}
This is a multivariate Gaussian. The factors in square brackets are the natural parameters of the Gaussian: the \emph{natural mean} (precision times mean); and the \emph{precision} (inverse covariance). For convenience, we have defined zero and first-order stats, $n_{s}^i$ and $\fvec_{s}^i$; as well as $\avec_s$ and $\Bmat_s$, which represent the data-dependent parts of the natural mean and precision. 

The language-independent posterior covariance is:
\begin{align}
\Cmat_s &= (\Wmat+\Bmat_s)^{-1}
\end{align}
The language-dependent posterior mean is:
\begin{align}
\begin{split}
\muvec_{\ell s} &= \Cmat_s(\Wmat\mvec_\ell + \avec_s) 
\end{split}
\end{align}
We shall later need the posterior expectations:
\begin{align}
\expv{\xvec}{Q_{\ell s}} &= \muvec_{\ell s} 
\end{align} 
and for some symmetric matrix $\Mmat$:
\begin{align}
\expv{\xvec'\Mmat\xvec}{Q_{\ell s}} &= \expv{\trace(\xvec\xvec'\Mmat)}{Q_{\ell s}}
= \trace\Bigl[\expv{\xvec\xvec'}{}\Mmat\Bigr]
= \trace\Bigl[\bigl(\Cmat_{s} + \muvec_{\ell s}\muvec_{\ell s}\bigr)\Mmat\Bigr] 
\end{align} 

\subsection{Parameter updates}
To learn the model parameters, we can alternate E and M steps. The E-step is computing the i-vector posterior. The M-step follows. To update $\mvec_\ell$, we need to maximize:
\begin{align}
\begin{split}
\sum_{s\in\Sset_\ell} \LB_{\ell s} &= \sum_{s\in\Sset_\ell} \expv{\log \ND(\xvec\mid\mvec_\ell,\Wmat^{-1})}{Q_{\ell s}} +\const\\
&= \sum_{s\in\Sset_\ell}  -\frac12\mvec'_\ell\Wmat\mvec_\ell + \mvec'_\ell\Wmat\muvec_{\ell s} +\const\\
\end{split}
\end{align}
which is maximized, independently of $\Wmat$, at:
\begin{align}
\mvec_\ell &= \bar\muvec_\ell = \frac1{\lvert \Sset_\ell \rvert}\sum_{s\in\Sset_\ell} \muvec_{\ell s} 
\end{align}
To update $\Wmat$, given the $\bar\muvec_\ell$, we need to maximize:
\begin{align}
\begin{split}
&=\sum_{\ell}\sum_{s\in\Sset_\ell} \LB_{\ell s}\\ 
&= \sum_{\ell,s} \expv{\log \ND(\xvec\mid\bar\muvec_\ell,\Wmat^{-1})}{Q_{\ell s}} +\const\\
&= \sum_{\ell,s} \frac12\logdetm{\Wmat} -\frac12\bar\muvec_\ell'\Wmat\bar\muvec_\ell 
+ \frac12\bar\muvec'_\ell\Wmat\muvec_{\ell s} 
+ \frac12\muvec'_{s}\Wmat\bar\muvec_\ell
- \frac12 \trace\bigl[\Wmat(\Cmat_{s}+\muvec_{\ell s}\muvec'_{\ell s})\bigr]+\const\\
&= \sum_{\ell,s} \frac12\logdetm{\Wmat}
-\frac12\trace\bigl[\Wmat(
\bar\muvec_\ell\bar\muvec'_\ell
-\bar\muvec_\ell\muvec'_{s}
-\muvec_{\ell s}\bar\muvec'_\ell
+\Cmat_{s}
+\muvec_{\ell s}\muvec'_{\ell s}
)\bigr]
\end{split}
\end{align}
which is maximized at:
\begin{align}
\begin{split}
\Wmat^{-1} &= \frac1N\sum_{\ell,s} \bar\muvec_\ell\bar\muvec'_\ell
-\bar\muvec_\ell\muvec'_{\ell s}
-\muvec_{\ell s}\bar\muvec'_\ell
+\Cmat_{s}
+\muvec_{\ell s}\muvec'_{\ell s} \\
&= \frac1N\sum_{\ell,s} 
\Cmat_{s}
+\bar\muvec_\ell(\bar\muvec_\ell-\muvec_{\ell s})'
+\muvec_{\ell s}(\muvec_{\ell s}-\bar\muvec_\ell)' \\
&= \frac1N\sum_{s=1}^N 
\Cmat_{s}
+\frac1N\sum_\ell\sum_{s\in\Sset_\ell}(\muvec_{\ell s}-\bar\muvec_\ell)(\muvec_{\ell s}-\bar\muvec_\ell)' \\
\end{split}
\end{align}
where $N$ is the total number of segments.\footnote{The middle line simplifies to the last because the second term is zero and the last term can be symmetrized by viewing mean subtraction as multiplication by the idempotent centering matrix.}

\subsection{Language scores}
We can form language scores (approximate log-likelihoods) by evaluating lower bound for each $\ell$, while omitting any terms independent of $\ell$. For a to-be-scored speech segment $s$, we compute separately for each language $\ell$, the lower bound: 
\begin{align}
\LB_{\ell s} &= \expv{\log\frac{\ND(\xvec\mid\mvec_\ell,\Wmat^{-1})}{Q_{\ell s}(\xvec)}
+\sum_{t=1}^{T_s} \sum_{i=1}^N q_{st}^i \log\frac
{w_i\ND(\phivec_{st}^i\mid\Tmat_i\xvec,\Imat)}
{q_{st}^i} }{Q_{\ell s}} 
\end{align}
We can simplify this expression by omitting any terms independent of $\ell$.\footnote{Note in particular, that the entropy term for $Q_{\ell s}$ is language-independent, because the entropy depends only on the covariance, not the mean.} 
\begin{align}
\begin{split}
\LB_{\ell s} &= \expv{
\log\ND(\xvec\mid\mvec_\ell,\Wmat^{-1})
+\sum_{t,i} q_{st}^i \log\ND(\phivec_{st}^i\mid\Tmat_i\xvec,\Imat)
}{Q_{\ell s}} +\const \\
&=-\frac12\mvec_\ell'\Wmat\mvec_\ell + \expv{
-\frac12\xvec'(\Wmat+\Bmat_s)\xvec +\xvec'(\Wmat\mvec_\ell
 + \avec_s) 
}{Q_{\ell s}} + \const\\
&=-\frac12\mvec_\ell'\Wmat\mvec_\ell 
-\frac12\trace\Bigl[\Cmat_s^{-1}(\Cmat_s+\muvec_{\ell s}\muvec'_{\ell s})\Bigr]
+ \muvec'_{\ell s}\Cmat_s^{-1}\muvec_{\ell s} + \const\\
&=-\frac12\mvec_\ell'\Wmat\mvec_\ell 
+ \frac12\muvec'_{\ell s}\Cmat_s^{-1}\muvec_{\ell s} + \const\\
&=-\frac12\mvec_\ell'\Wmat\mvec_\ell 
+ \frac12(\Wmat\mvec_\ell+\avec_s)'\Cmat_s(\Wmat\mvec_\ell+\avec_s) + \const\\
&=-\frac12\mvec_\ell'\Wmat\mvec_\ell 
+ \frac12\mvec_\ell'\Wmat\Cmat_s\Wmat\mvec_\ell 
+ \mvec'_\ell\Wmat\Cmat_s\avec_s + \const
\end{split}
\end{align}
So let's drop the constant terms and define the language score as:
\begin{align}
\sigma_{\ell s} &= -\frac12\mvec_\ell'(\Wmat-\Wmat\Cmat_s\Wmat)\mvec_\ell 
+ \mvec'_\ell\Wmat\Cmat_s\avec_s
\end{align}
To examine the behaviour of this score, keep in mind $\Cmat_s=(\Wmat+\Bmat_s)^{-1}$; and that $\avec_s$ and $\Bmat_s$ are zero at $T_s=0$ and keep increasing with $T_s$. At $T_s=0$, we get the nice effect $\sigma_{\ell s}=0$. Conversely, for large $T_s$, we find $\Wmat\Cmat_s\Wmat$ eventually vanishes, while $\Cmat_s\avec_s$ converges to $\tilde\muvec_s$, the classical i-vector and the score reduces to that given by the by the stand-alone linear Gaussian back-end:
\begin{align}
\sigma_{\ell s}\rvert_{T_s\gg1} \approx -\frac12\mvec_\ell'\Wmat\mvec_\ell 
+ \mvec'_\ell\Wmat\tilde\muvec_s
\end{align}

\subsubsection{Practical scoring}
The above scoring recipe is expressed in terms of $\avec_s=\sum_i \Tmat'_i\fvec_s^i$ and $\Bmat_s=\sum_i\Tmat'_i\Tmat_i n_s^i$, which are in turn obtained from the first and zero-order stats, $\fvec^i_s$ and $n^i_s$. We can therefore apply this recipe without explicitly going via i-vectors. 

It is however possible (and perhaps preferable) to instead apply this recipe using already extracted i-vectors. The i-vectors are \emph{much} smaller than the first-order stats and therefore much easier to work with on disk and in memory. The classical i-vector is:
\begin{align}
\tilde\muvec_s &= (\Imat+\Bmat_s)^{-1}\avec_s
\end{align}
We can therefore recover $\avec_s$ from the i-vector:
\begin{align}
\avec_s&= (\Imat+\Bmat_s)\tilde\muvec_s
\end{align}
Of course, we also need $\Bmat_s$. As long as we have $\Tmat_i$ available, then $\Bmat_s$ can be computed via the zero-order stats, $n_s^i$, which are only moderately larger then the i-vectors and can be conveniently stored alongside them.

\subsection{CPF Scoring}
\def\Emat{\mathbf{E}}
We can compare the above scoring recipe to Cumani-Plchot-F\'er (CPF) scoring~\cite{CPF}. In the CPF recipe, the i-vector is also integrated out, but the classical i-vector posterior is used instead of the $Q_{\ell s}$ of section~\ref{sec:Qls}. The difference between the classical posterior and $Q_{\ell s}$ is that the classical one uses a language-independent, standard normal prior. We can therefore expect $Q_{\ell s}$ to be closer to the true posterior. 

Denoting the classical posterior precision as $\Emat_s=\Imat+\Bmat_s$, the classical, language independent i-vector (posterior mean) is $\tilde\muvec_s=\Emat_s^{-1}\avec_s$. The CPF score is:\footnote{We use the identity: $(\Wmat^{-1}+\Emat_s^{-1})^{-1}=\Wmat(\Wmat+\Emat_s)^{-1}\Emat_s$. If this seems surprising, just look at the scalar case: $\frac1{w^{-1}+e^{-1}}=\frac{we}{w+e}$.}
\begin{align}
\begin{split}
&-\frac12(\tilde\muvec_s-\mvec_\ell)'(\Wmat^{-1}+\Emat_s^{-1})^{-1}(\tilde\muvec_s-\mvec_\ell) \\
&= -\frac12\mvec_\ell'(\Wmat^{-1}+\Emat_s^{-1})^{-1}\mvec_\ell 
+ \mvec'_\ell(\Wmat^{-1}+\Emat_s^{-1})^{-1}\tilde\muvec_s +\const \\
&= -\frac12\mvec_\ell'\Wmat(\Wmat+\Emat_s)^{-1}\Emat_s\mvec_\ell 
+ \mvec'_\ell\Wmat(\Wmat+\Emat_s)^{-1}\Emat_s\tilde\muvec_s +\const \\
&= -\frac12\mvec_\ell'\Wmat(\Wmat+\Emat_s)^{-1}\Emat_s\mvec_\ell 
+ \mvec'_\ell\Wmat(\Wmat+\Emat_s)^{-1}\avec_s +\const 
\end{split}
\end{align}
Defining $\tilde\Cmat_s=(\Wmat+\Emat_s)^{-1}=(\Wmat+\Imat+\Bmat_s)^{-1}$ in analogy to $\Cmat_s=(\Wmat+\Bmat_s)^{-1}$ and dumping constant terms, the CPF score is:
\begin{align}
\tilde\sigma_{\ell s}&= -\frac12\mvec_\ell'\Wmat\tilde\Cmat_s\Emat_s\mvec_\ell 
+ \mvec'_\ell\Wmat\tilde\Cmat_s\avec_s
\end{align}
This score behaves like $\sigma_{\ell s}$ for large $T_s$ and also converges to the standalone linear Gaussian back-end. But for very small $T_s$, the behaviour is different---in particular this score does not become independent of language at $T_s=0$:
\begin{align}
\tilde\sigma_{\ell s}\rvert_{T_s=0} &= -\frac12\mvec_\ell'\Wmat(\Wmat+\Imat)^{-1}\mvec_\ell 
\end{align}

\end{document}